\documentclass[fleqn,10pt]{wlscirep}

\usepackage{multirow}
\usepackage{lineno}
\newcommand{\deltaf}{\ensuremath{\Delta\text{F1}}}

\title{Label-efficient underwater image classification with logistic regression on frozen foundation model embeddings}

\author[1,2,4,5,7,*]{Thomas Manuel Rost}
\author[3]{Martina Figlia}
\author[1,2]{F. Morgado-Dias}
\author[1,4,5,6]{Marko Radeta}

\affil[1]{University of Madeira, Faculty of Exact Sciences and Engineering, Funchal, Portugal}
\affil[2]{Interactive Technologies Institute (ITI/LARSyS), ARDITI, Funchal, Portugal}
\affil[3]{Institute of Cognitive Science, University of Osnabrueck, Osnabrueck, Germany}
\affil[4]{Wave Labs, Faculty of Arts and Humanities, University of Madeira, Portugal}
\affil[5]{MARE - Marine and Environmental Sciences Centre, ARNET - Aquatic Research Network, Regional Agency for Research, Technology and Innovation Development (ARDITI), Funchal, Portugal}
\affil[6]{Department of Astronomy, Faculty of Mathematics, University of Belgrade, Serbia}
\affil[7]{OKEANOS, University of the Azores, Faial, Portugal}
\affil[*]{Correspondence: thomas.rost@mare.arditi.pt}

\keywords{foundation models, logistic regression, underwater image classification, DINOv3, label efficiency, cost-efficient marine science}

\begin{document}

\begin{abstract}
Underwater image classification is constrained by the cost of annotation and by the computational and methodological requirements of task-specific model development. We investigate whether frozen general-purpose foundation-model embeddings can reduce these requirements by extracting DINOv3 ViT-B/16 embeddings and training only a logistic regression classifier on the AQUA20 benchmark. We evaluate the approach across a range of annotation budgets, a repeated 80\% training-subsample evaluation, and a full-training refit. With only 13 labelled images per category, corresponding to approximately 4\% of the benchmark's official training partition, mean macro F1 reaches 81.8\%; with 144 images per category it reaches 88.5\%, compared to the published fully supervised ConvNeXt point estimate of 88.9\% obtained with the complete training set (benchmark results reported without run-to-run variability). Using all official training labels, macro F1 reaches 91.5\% (bootstrap 95\% CI: 89.0--93.7\%).

Sensitivity analyses show that the main findings remain stable across ordinary downstream implementation choices, and persist after removing duplicate and near-duplicate test images identified in an audit of the official split. 
Preliminary evaluation on a second dataset suggests that overall performance level and the shape of the label efficiency curve are not unique to the AQUA20 dataset. 
Because the DINOv3 backbone remains frozen and only the downstream classifier is fitted, the approach avoids task-specific neural-network training and can be executed on commodity hardware. These findings establish linear classification on frozen foundation-model embeddings as a practical baseline for label-efficient underwater image classification.
\end{abstract}

\flushbottom
\maketitle
\thispagestyle{empty}

\section*{Introduction}\label{sec:introduction}

Visual classification is a task encountered across many areas of marine science: monitoring biodiversity on coral reefs, identifying fish species in commercial catch assessments, recognising individual whales from fluke photographs, classifying plankton in high-throughput imaging systems, and analysing video transects from deep-sea exploration platforms such as the Azor drift-cam~\cite{dominguezcarrio2021azordriftcam,goodwin2022unlocking,radeta2022deeplearning_oceans}. Researchers often turn towards computer vision to aid in the processing of such data, but conventional approaches typically require a task-specific model adapted to a dataset that closely matches the target conditions; changes in depth, lighting, turbidity, camera angle, equipment, or geographic location can each cause substantial degradation in accuracy~\cite{saleh2022survey,mittal2023underwater_survey}. Converting the growing volume of underwater visual data, driven by remotely operated vehicles, autonomous platforms, and fixed observatories, into species- or individual-level observations therefore remains a critical bottleneck: taxonomic identification requires domain expertise, annotation can be difficult even for specialists, and new imaging conditions frequently demand additional labelled data. In practice, much of the collected material is still analysed manually or with only limited computational assistance~\cite{benthicnet2025,bigham2026}.

Where underwater visual classification is automated, the dominant approach remains task-specific supervised deep learning: models such as ConvNeXt and Vision Transformers are commonly pretrained and subsequently fine-tuned end-to-end on labelled underwater datasets, and have achieved strong results on benchmarks including AQUA20~\cite{fuad2026aqua20}, a 20-category underwater classification benchmark. However, assembling the required datasets and subsequently training systems is expensive. End-to-end model development requires computational resources, machine-learning expertise, and numerous implementation decisions concerning model architecture, optimisation, data augmentation, hyperparameter selection, and checkpoint selection. The resulting models can remain tightly coupled to their training distribution (limiting reusability), while careful dataset preparation and quality control are necessary to obtain reliable results. Together, these requirements continue to limit the routine adoption of deep learning methods in many marine-science applications.

Meanwhile, self-supervised vision foundation models such as DINOv2~\cite{oquab2024dinov2} and DINOv3~\cite{dinov32025} have demonstrated that general-purpose visual representations, learned without task-specific class labels, can match or approach supervised performance across a wide range of downstream tasks in combination with simple classifiers. Rather than learning a new visual representation for each application, images can be projected into a fixed embedding space using a frozen foundation model, after which only a lightweight classifier is fitted for the target task. Because these representations are learned during large-scale pretraining, these systems may also require considerably fewer labelled examples to reach useful performance than end-to-end approaches. Recent work has shown that frozen DINOv3 embeddings separate terrestrial camera-trap imagery into species-level clusters without any class labels, approaching supervised clustering performance, and that this general-purpose representation outperforms the biology-specific BioCLIP~2 embedding on the same task~\cite{markoff2026vit_clustering}. In marine computer vision, where used, foundation-model approaches have generally relied on marine-specific representations or task-specific adaptation~\cite{alawode2025aquaticclip,coral_lora_2025,diveseg2025}. This raises a natural question: if general-purpose foundation-model embeddings already capture sufficient morphological structure for category-level discrimination, can a simple classifier on these frozen features achieve competitive classification accuracy in the underwater domain?

In this work, we investigate exactly this question. We extract frozen DINOv3 embeddings for all images in the AQUA20 underwater visual-classification benchmark and evaluate logistic regression across a range of labelling budgets, from as few as 1 to 144 nominal labelled examples per class, as well as a repeated stratified 80\% training-subsample regime and a final refit using 100\% of the official training labels. We find that logistic regression on frozen, non-fine-tuned foundation-model embeddings reaches performance comparable to the published fully supervised ConvNeXt point estimate while using substantially fewer labelled examples, indicating that the frozen DINOv3 embedding space provides a linearly decodable representation for underwater visual categories without underwater-specific adaptation. To further assess performance, we additionally examine per-category behaviour, computational requirements, sensitivity to downstream implementation choices, common failure modes, and the integrity of the official benchmark split through duplicate and label-quality audits. Full quantitative results, including the label-efficiency curve and per-class comparison, are reported in the Results. Because the foundation model used here was not adapted to underwater imagery, domain-adapted foundation models may further improve performance, although this remains to be tested. Furthermore, because our pipeline relies on a general-purpose model rather than a domain-specific and task-optimised one, it should in principle transfer to other environments and datasets; a preliminary external validation is reported in the Supplementary Information, and broader evaluation remains for future work.

\section*{Related Work}
\label{sec:related}

\subsection*{Computer Vision in Marine Science}
The rapid growth of underwater image collections has driven substantial interest in automated image analysis based on deep learning. Recent surveys describe applications spanning species classification, habitat mapping, coral reef monitoring, and object detection, with end-to-end neural networks becoming a widely used approach across these tasks~\cite{saleh2022survey,mittal2023underwater_survey,radeta2022deeplearning_oceans}. A wide variety of task-specific architectures have been evaluated, including convolutional neural networks and vision transformers, often through dedicated training or adaptation for individual datasets and applications~\cite{fuad2026aqua20,mehrab2024fishvista}. Large-scale resources such as Fish-Vista have further advanced supervised classification by providing benchmarks spanning thousands of species~\cite{mehrab2024fishvista}. AQUA20, the benchmark we adopt here, compares thirteen architectures under realistic underwater degradation conditions and reports its strongest performance for an ImageNet-pretrained ConvNeXt model fine-tuned end-to-end on the benchmark~\cite{fuad2026aqua20}.

\subsection*{Foundation Models}

More recently, large-scale vision foundation models have emerged as a powerful alternative to task-specific representation learning. Beginning with DINO~\cite{caron2021dino}, the Vision Transformer~\cite{dosovitskiy2021vit} was shown to learn semantically rich visual representations through self-distillation without any labels. Its successors DINOv2~\cite{oquab2024dinov2} and DINOv3~\cite{dinov32025} scale this approach and produce general-purpose representations that transfer across a wide range of downstream tasks via lightweight classifiers, with no fine-tuning of the backbone. DINOv3 is available in many sizes, including the 86-million parameter version used in this study which is itself distilled from a self-supervised teacher trained on roughly 1.7 billion web images. Instead of learning visual features separately for each application, these pretrained representations can be reused by fitting lightweight downstream models, substantially reducing the need for task-specific optimisation while maintaining competitive performance across many domains. These frozen representations have proven effective in ecological domains: DINOv2 features have transferred to plant-species classification and agricultural segmentation across sensor modalities~\cite{gustineli2024plantclef,picon2026agricultural}, while frozen DINOv3 embeddings yield near-perfect species-level clustering of terrestrial camera-trap imagery, outperforming specialised vision--language models such as BioCLIP~\cite{markoff2026vit_clustering}. As a result, foundation models have increasingly shifted the focus from training deep representations to reusing them.

\subsection*{Foundation Models in Marine Science}
In marine science, foundation-model-based work has followed several different routes. AquaticCLIP develops a marine-specific multimodal representation from underwater image-text pairs~\cite{alawode2025aquaticclip}, while coral reef monitoring with DINOv2 uses LoRA-based fine-tuning of a pretrained backbone~\cite{coral_lora_2025}. DiveSeg similarly introduces task-specific adaptation modules and argues that direct frozen transfer is not straightforward because of the underwater domain gap~\cite{diveseg2025}. In a different modality, a related result shows that a linear classifier on frozen audio embeddings can recognise underwater ship types, indicating that simple decoders on frozen representations can be effective for underwater sensing beyond the visual domain~\cite{hummel2026linear}. However, none of these studies establishes whether a fully frozen, general-purpose visual representation combined with a lightweight classifier can provide a practical workflow for underwater image classification.

After the preprint~\cite{rost2026label} of this work became available, Saleh and Rahimi Azghadi independently evaluated frozen DINOv2 representations with linear probing and several adaptation strategies across five marine datasets, including cross-habitat and few-shot settings~\cite{saleh2026howmanylabels}. Their results provide complementary evidence that frozen general-purpose representations can remain effective across marine datasets and low-label regimes.

\subsection*{Data Quantity and Quality}
The success of machine learning methods in marine science depends strongly on the quantity and quality of the available annotations. Developing reliable deep learning models typically requires both sufficient volumes of labelled data and consistent expert annotation, making dataset construction one of the most time-consuming and resource-intensive stages of many marine imaging projects~\cite{benthicnet2025,bigham2026}. Consequently, considerable effort has been devoted to improving annotation workflows through better datasets, semi-automated annotation, and machine-learning-assisted labelling. While these advances reduce the effort required to construct annotated datasets, the literature places considerably greater emphasis on improving predictive performance than on reducing the annotation requirements of the modelling approach itself.

\subsection*{Our Contribution}
This work builds on these developments by evaluating frozen foundation model embeddings from both practical and predictive perspectives. Rather than introducing a new architecture, training a marine-specific representation, or adapting an existing foundation model, we investigate whether a frozen general-purpose representation combined with a lightweight classifier provides a practical workflow for label-efficient underwater image classification. We select DINOv3 because it provides particularly strong evidence of species-level structure in frozen general-purpose embeddings on the closest available analogue task, indicating that its embeddings might carry potential for label efficiency, while its 86-million-parameter ViT-B/16 checkpoint remains suitable for embedding extraction on commodity hardware~\cite{dinov32025,markoff2026vit_clustering}. To our knowledge, this is the first systematic evaluation of fully frozen, general-purpose vision foundation model embeddings with a simple classifier for label-efficient underwater image classification on AQUA20. We assess predictive performance alongside annotation efficiency, computational requirements, downstream robustness, and per-category behaviour. Additionally, we examine the integrity of the benchmark itself through duplicate and label-quality audits, providing context for the interpretation of the reported results.

\section*{Methods}

\subsection*{Data and Preprocessing}

\begin{figure*}[p]
    \centering
    \newsavebox{\AQUAgrid}
    \sbox{\AQUAgrid}{\includegraphics[width=0.39\textwidth]{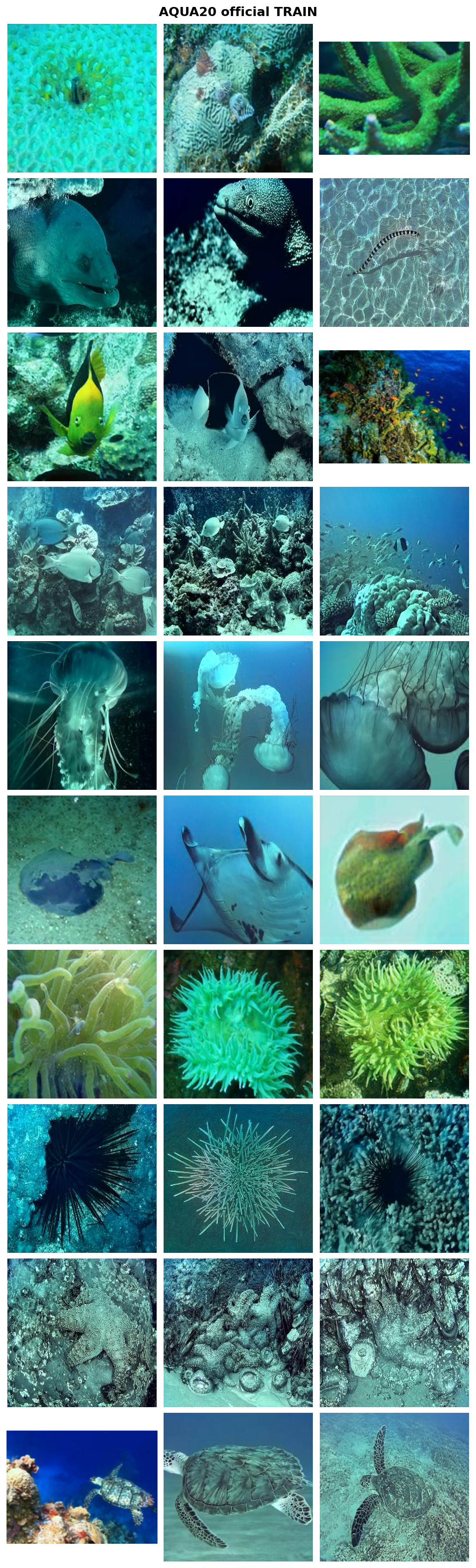}}
    
    \setlength{\tabcolsep}{2pt}
    \begin{tabular}{c c c}
        \usebox{\AQUAgrid} & 
\begin{minipage}[b][\ht\AQUAgrid][s]{0.14\textwidth}
            \centering
            \scriptsize
            \vspace{4em} 
            
            \textbf{Coral} \vfill
            \textbf{Eel} \vfill
            \textbf{Fish} \vfill
            \textbf{FishInGroups} \vfill
            \textbf{Jellyfish} \vfill
            \textbf{Rayfish} \vfill
            \textbf{SeaAnemone} \vfill
            \textbf{SeaUrchin} \vfill
            \textbf{Starfish} \vfill
            \textbf{Turtle} 
            
            \vspace{2em} 
        \end{minipage} &
        \includegraphics[width=0.39\textwidth]{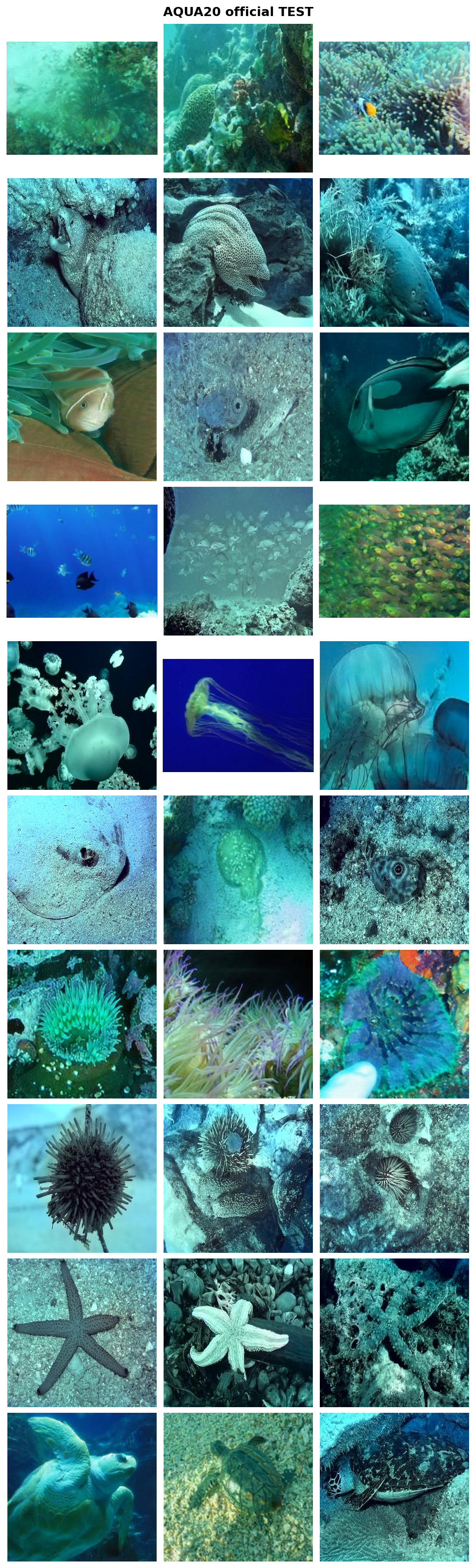}
    \end{tabular}

    \caption{Randomly selected sample images from ten of the AQUA20 visual categories from the official train (left) and test (right) splits, illustrating the diversity of morphology, colouration, and imaging conditions across the dataset.}
    \label{fig:dataset-overview}
\end{figure*}

We evaluate on the AQUA20 benchmark dataset~\cite{fuad2026aqua20}, which comprises 8,171 underwater images across 20 marine image categories including fish, coral formations, crustaceans, cephalopods, marine mammals, and human divers. The dataset combines images from the EUVP (Enhancement of Underwater Visual Perception) dataset and other publicly available underwater image sources~\cite{fuad2026aqua20}. It is pre-split into 6,559 training and 1,612 test images and was specifically curated to reflect environmental challenges commonly encountered in underwater imaging, including turbidity, low illumination, colour distortion, and partial occlusion. AQUA20 also exhibits severe class imbalance, with classes such as Fish containing 2,200 training images, whereas MarineDolphin and Octopus contain only 20 training images each. These characteristics make AQUA20 a suitable benchmark for evaluating both predictive performance and label efficiency under realistic underwater imaging conditions. The class distribution is summarised in Table~\ref{tab:class_counts}, while Figure~\ref{fig:dataset-overview} illustrates representative images from the official training and test partitions. Since AQUA20 images are already cropped to individual organisms, no additional segmentation was applied.
Throughout this work, category names are rendered in PascalCase (e.g., MarineDolphin, SeaAnemone, FishInGroups) for consistency.

In addition to conducting the classification experiments, we audited the official AQUA20 train/test split for duplicate images and potential annotation inconsistencies. We screened the dataset using three complementary signatures: SHA-256 hashes of the raw file bytes, SHA-256 hashes of the decoded RGB pixel arrays, and perceptual hashing (pHash and dHash) using a Hamming-distance threshold of four while requiring both perceptual hashes to agree~\cite{buchner_imagehash}. Candidate duplicate clusters spanning the official training and test partitions were subsequently inspected visually. Note that in order to preserve comparability with the published AQUA20 benchmark, all classification experiments were performed on the unchanged official splits without modification, we report the outcomes of the audit in the discussion and limitations. The complete audit methodology is provided in Supplementary Note 1; the results of the audit are reported in the Results section and discussed in the Limitations.

\begin{table}[htpb]
\caption{Official AQUA20 dataset class distribution, detailing the number of training, test, and total samples per category. Counts are taken directly from the dataset description.}
\label{tab:class_counts}
\small
\begin{tabular*}{\columnwidth}{@{\extracolsep{\fill}} rlrrr @{}}
\toprule
No. & Class & Train & Test & Total \\
\midrule
1 & Coral & 1562 & 348 & 1910 \\
2 & Crab & 43 & 11 & 54 \\
3 & Diver & 51 & 13 & 64 \\
4 & Eel & 160 & 41 & 201 \\
5 & Fish & 2200 & 538 & 2738 \\
6 & FishInGroups & 272 & 72 & 344 \\
7 & Flatworm & 50 & 13 & 63 \\
8 & Jellyfish & 97 & 25 & 122 \\
9 & MarineDolphin & 20 & 10 & 30 \\
10 & Octopus & 20 & 10 & 30 \\
11 & Rayfish & 382 & 95 & 477 \\
12 & SeaAnemone & 890 & 221 & 1111 \\
13 & SeaCucumber & 35 & 10 & 45 \\
14 & SeaSlug & 79 & 20 & 99 \\
15 & SeaUrchin & 115 & 29 & 144 \\
16 & Shark & 71 & 19 & 90 \\
17 & Shrimp & 22 & 11 & 33 \\
18 & Squid & 24 & 10 & 34 \\
19 & Starfish & 166 & 40 & 206 \\
20 & Turtle & 300 & 76 & 376 \\
\midrule
& \textbf{Total} & \textbf{6559} & \textbf{1612} & \textbf{8171} \\
\bottomrule
\end{tabular*}
\end{table}

\subsection*{Embedding Extraction}

All images are converted to RGB and processed using the AutoImageProcessor associated with the facebook/dinov3-vitb16-pretrain-lvd1689m checkpoint. For the main experiments, images are resized to $384 \times 384$ pixels and transformed using the checkpoint's default rescaling and normalisation parameters; no data augmentation is applied. This resolution is divisible by the model's 16-pixel patch size and retains substantially more spatial detail than the processor's standard $224 \times 224$ input. We compare three resolutions ($224$, $384$ , $512$) in an input-resolution ablation in the Supplementary Information.
The processed images are then embedded using the frozen 86\,M DINOv3 ViT-B/16 checkpoint \texttt{facebook/dinov3-vitb16-pretrain-lvd1689m}~\cite{dinov32025}. The model contains 12 transformer blocks, uses a patch size of 16 pixels, and produces 768-dimensional token representations. We extract the final-layer CLS token, which serves as the global image representation, and apply L2 normalisation, yielding a single 768-dimensional float32 embedding for each image. The model additionally produces four register tokens that absorb high-norm artefacts~\cite{darcet2024registers} and spatially localised patch tokens. Embeddings are extracted in batches of four images.

While the register and patch tokens are discarded in the main experiments, we provide an alternative pooling strategy using the patch tokens in the Supplementary Information (pooling ablation). No dimensionality reduction is applied; the full 768-dimensional embeddings are used directly. A PCA ablation is reported in the Supplementary Information (dimensionality reduction). Embedding extraction was performed with the model in evaluation mode under inference-only execution. Repeated extraction on the Apple MPS backend nevertheless produced non-bit-identical embeddings at the level of approximately $10^{-6}$ after L2 normalisation. All reported experiments therefore use a single archived embedding cache.
Re-extracting embeddings and refitting the classifier reproduced the reported macro F1 to three decimal places in both the 100\% refit and the repeated 80/20 evaluation, confirming that the noise floor does not affect downstream results.

\begin{figure*}[t]
    \centering
    \includegraphics[width=0.90\textwidth]{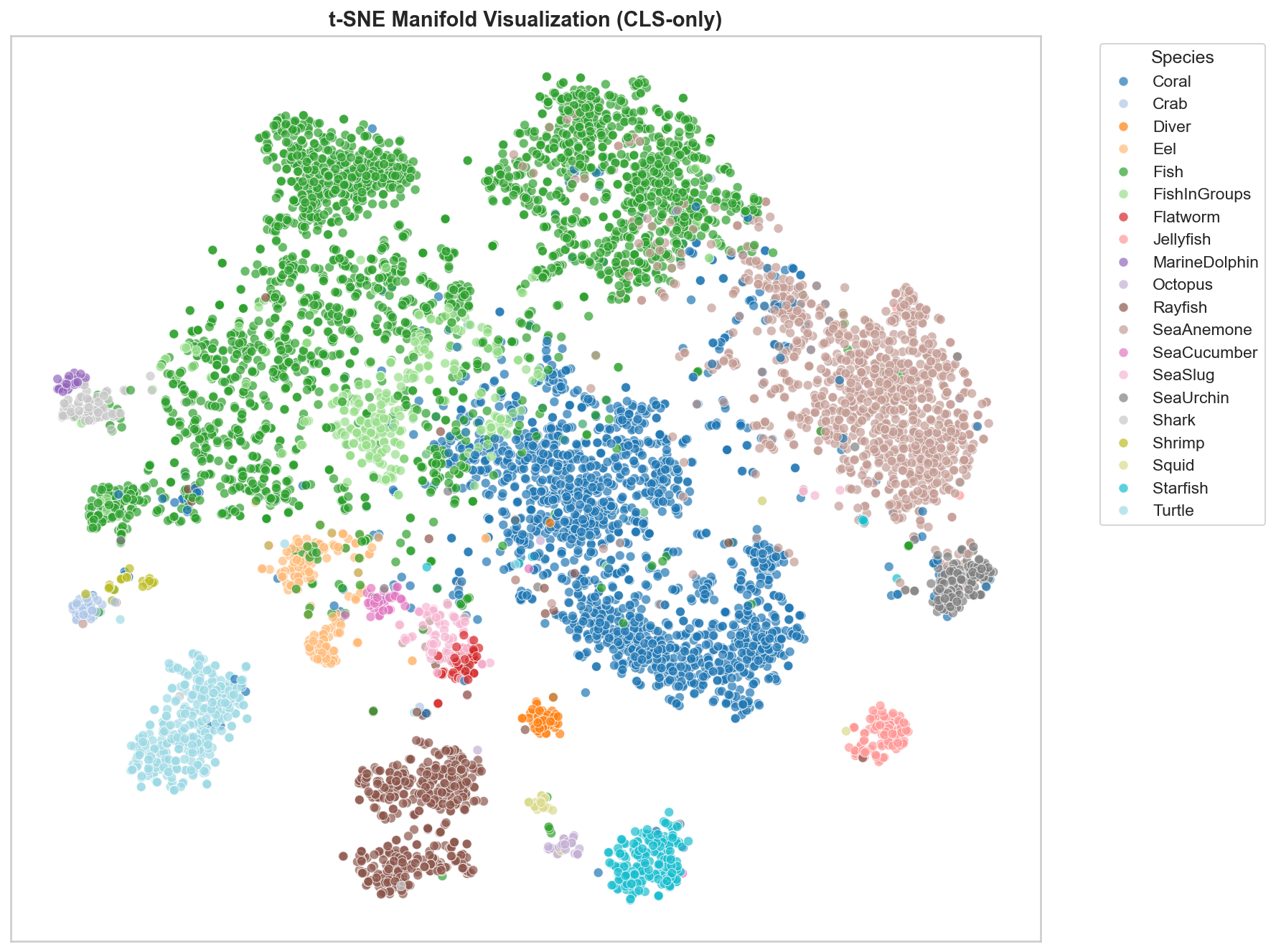}
    \caption{t-SNE projection of the frozen DINOv3 embeddings for the complete AQUA20 dataset, coloured by ground-truth category. Several categories form compact, well-separated clusters despite the absence of underwater-specific training. The projection is shown for qualitative illustration only; the axes have no intrinsic meaning and global distances are not preserved.}
    \label{fig:tsne}
\end{figure*}

To qualitatively assess the structure of the frozen embedding space, Figure~\ref{fig:tsne} presents a t-distributed Stochastic Neighbor Embedding (t-SNE) visualisation of the extracted DINOv3 embeddings, coloured by ground-truth category. t-SNE projects the 768-dimensional embedding space into two dimensions while preserving local neighbourhood relationships, allowing nearby points in the projection to correspond to visually similar images in the original feature space. The resulting clusters provide qualitative evidence that the frozen foundation model captures substantial category-discriminative structure without underwater-specific training. Note that with t-SNE visualisations, global distances and the coordinate axes themselves should not be interpreted quantitatively.

\paragraph{Reproducibility.}

To facilitate reproduction of the reported downstream experiments, we archive the extracted CLS-token and patch-mean embedding caches for AQUA20 and Sea Animals together with a machine-readable metadata file recording the model and processor configurations, requested image resolution, software versions, hardware information, numerical precision, and random seeds used throughout the study. The complete embedding-extraction and downstream-analysis code is released on GitHub, while the embeddings are archived on Zenodo.

All components of the pipeline described in this work are publicly available as open source. 
The foundation model is unmodified and the classifier is a standard scikit-learn implementation. No proprietary software or data is required at any stage.

\subsection*{Classification}

We train a logistic regression classifier on the frozen DINOv3 embeddings using the L-BFGS solver. All final experiments use balanced class weights, a maximum of 5{,}000 iterations, and a convergence tolerance of $10^{-8}$; the regularisation strength $C = 100$ was selected in a preliminary hyperparameter search. Solver settings were not searched, and their effect is verified post hoc in the Supplementary Information. Hyperparameters were selected once before the main evaluation using ten random stratified 80/20 splits of the official training partition with a seed sequence different from the main experiments. We evaluated $C \in \{0.1, 1, 10, 100, 1000\}$, after which the best-performing configuration was fixed for all subsequent experiments. The official test set was not used during hyperparameter selection. Supplementary analyses evaluate the sensitivity of the pipeline to regularisation strength, solver convergence, pooling strategy, and dimensionality reduction. We additionally compare logistic regression against classifiers with different inductive biases, including $k$-nearest neighbours, linear discriminant analysis, and Gaussian naive Bayes; logistic regression achieved the strongest overall performance, while $k$-nearest neighbours proved competitive.

\paragraph{Implementation sensitivity.}

To assess the robustness of the proposed workflow to common implementation choices, we additionally evaluate the effects of input resolution, classifier selection, regularisation strength, solver convergence, embedding pooling strategy, and dimensionality reduction. The input-resolution ablation compares the $384 \times 384$ resolution used in the main experiments with the standard $224 \times 224$ as well as a higher $512 \times 512$ input resolution in a strict train/validation setup, satisfying two constraints: classification performance as measured by macro F1 as well as runtime on commodity hardware. Unless otherwise stated, these analyses use the same frozen DINOv3 embeddings, training protocol, and evaluation procedure as the main experiments, with only the component under investigation being varied. Complete experimental settings and results are provided in the Supplementary Information.

\paragraph{Runtime measurement.}

Runtime measurements were obtained separately for embedding extraction and downstream classifier training. Embedding extraction was timed for the complete AQUA20 dataset using the frozen DINOv3 backbone in inference mode, while classifier runtimes were measured for the repeated 80/20 evaluation protocol. Reported runtimes exclude model download and dataset loading but include all image preprocessing, embedding extraction, and classifier fitting performed during the respective stage. Measurements were obtained on both an Apple M4 MacBook Air (16\,GB RAM) and an NVIDIA RTX A6000 workstation.

\subsection*{Experimental Design}

To evaluate the practical utility of frozen foundation model embeddings under different annotation scenarios, we perform three complementary experiments. First, we evaluate performance across a wide range of absolute per-class annotation budgets to quantify label efficiency. Second, we assess performance under a conventional repeated 80/20 train/validation protocol that reflects a typical machine learning workflow. Finally, after hyperparameter selection, we refit the classifier using 100\% of the available training data to estimate the maximum performance attainable for our pipeline under the official AQUA20 benchmark protocol.

All experiments use the official AQUA20 train/test partition. Hyperparameter selection, model fitting, and label-budget sampling are performed exclusively on the official training partition, while the official test set is reserved for final evaluation throughout. This preserves direct comparability with the published benchmark while preventing the use of test data during model and parameter selection.

\begin{figure*}[t]
  \centering
  \includegraphics[width=\textwidth]{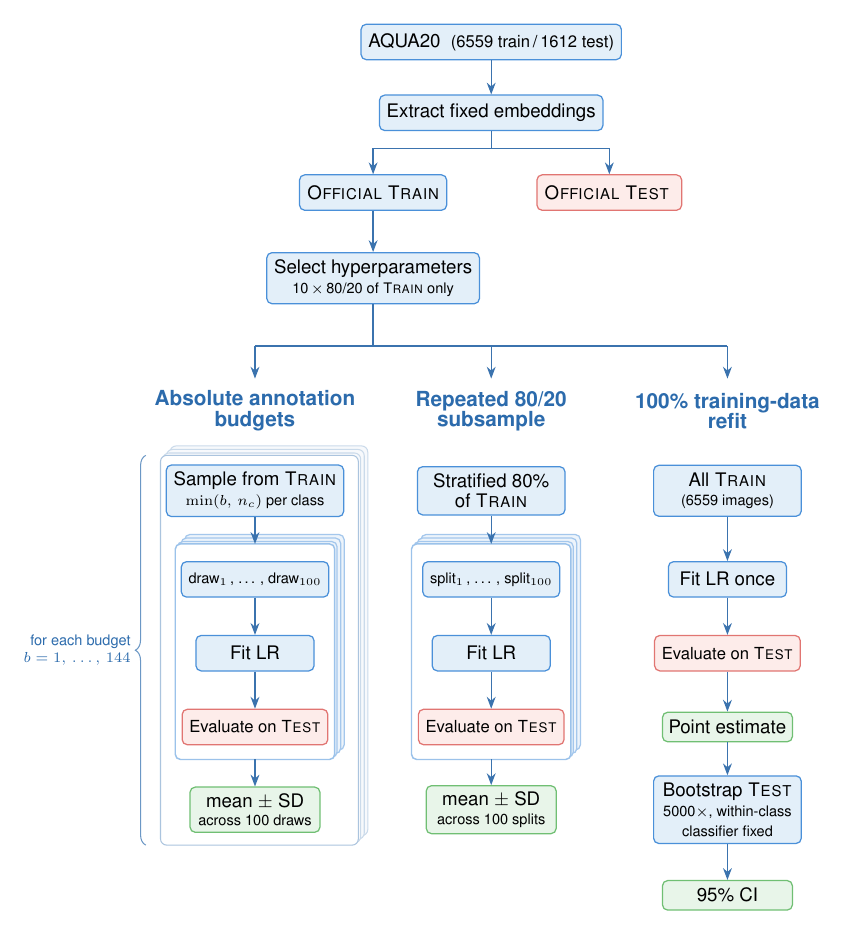}
  \caption{Overview of the overall design flow. Three separate experiments are run: various annotation budget, an 80/20 train-validation split and a 100\% refit. All experiments share hyperparameters, and the official AQUA20 test set is only used to calculate the final evaluation statistics. Stacks indicate several versions or iterations.}
  \label{fig:experimental_design}
\end{figure*}

An overview of the experimental set up can be found in Figure~\ref{fig:experimental_design}.

\paragraph{Annotation budgets.}

Our objective is to characterise the practical performance of frozen foundation model embeddings across a range of annotation budgets while providing the published ConvNeXt benchmark on AQUA20 as a reference. We therefore evaluate the proposed pipeline under three complementary annotation scenarios, spanning extreme label scarcity, a conventional machine learning workflow, and training on the complete official training partition.

\paragraph{Absolute annotation budgets.}

To quantify label efficiency, we evaluate absolute per-class annotation budgets of $b \in \{1, 2, 3, 5, 8, 13, 21, 34, 55, 89, 144\}$ labelled images. For each budget and each class $c$, we randomly sample $\min(b, n_c)$, where $n_c$ denotes the number of available training images for that class. Classes containing fewer than $b$ training images contribute all available examples and stay below their maximum allowed number of labels. Each annotation budget is evaluated over 100 independent random samples to quantify variability arising from the selection of labelled training images. The chosen budgets follow an approximately Fibonacci-spaced schedule, providing finer resolution in the extreme low-data regime, where small increases in annotation effort can substantially affect performance, while avoiding redundant evaluations once learning curves begin to plateau~\cite{meek2002learningcurve,viering2023shape}.

\paragraph{Repeated 80/20 training-subsample evaluation.}

To quantify sensitivity to the available training sample, we repeatedly partition the official training data into stratified 80/20 splits. Hyperparameters remain fixed throughout this evaluation. In each run, the classifier is fitted on the 80\% training subset; the remaining 20\% is not used for fitting, further model selection, or performance reporting. Performance is assessed on the official test set over 100 independent splits, quantifying the variability introduced by different training subsets. Unlike the original AQUA20 benchmark, which selects the best-performing checkpoint during end-to-end training, hyperparameter selection in our study is performed exclusively on separate validation splits drawn from the official training partition and is completed before these repeated-subsampling experiments.

\paragraph{100\% training-data refit.}

After hyperparameter selection, the classifier is refitted once using the complete official training partition. This experiment represents the practical deployment scenario in which all available labelled data are used for training and provides the maximum available data evaluation of the selected pipeline on the official AQUA20 train/test split. Because the training set and optimisation procedure are fixed, repeated fitting is deterministic and does not provide a meaningful estimate of uncertainty. 

Instead, uncertainty is quantified by class-stratified bootstrapping of the official test set. For each of 5{,}000 bootstrap replicates, test images are sampled with replacement within each class while preserving the original class frequencies, after which macro F1, weighted F1, and accuracy are recomputed. Percentile 95\% confidence intervals based on the 2.5th and 97.5th percentiles are reported.

\paragraph{External validation.}

To assess whether the observed behaviour extends beyond AQUA20, we additionally evaluate the proposed pipeline on the publicly available Sea Animals dataset. We follow the same experimental pipeline as for AQUA20, with one important exception: after identifying duplicate images and conflicting duplicate labels, these samples were removed to obtain a clean evaluation dataset. Because the Sea Animals dataset has not been formally introduced as a published benchmark, this experiment is intended as an external validation of the proposed workflow rather than as a direct benchmark comparison. Full details are provided in the Supplementary Information.
\paragraph{Reporting.}

Unless otherwise stated, all reported performance metrics are calculated on the held-out official AQUA20 test set. For the repeated 80/20 and absolute annotation-budget experiments, results are reported over 100 independent runs as mean $\pm$ one standard deviation. Because the official test set remains fixed throughout these experiments, the reported standard deviations reflect variability arising from the sampled training data rather than from the test set itself. For the 100\% training-data refit, a single point estimate is reported together with the bootstrap confidence intervals described above. Unless otherwise stated, we report macro F1, support-weighted F1, and accuracy. Owing to rounding, reported differences may not correspond exactly to differences computed from the rounded values shown.
Because AQUA20 reports only a single point estimate for the ConvNeXt baseline, we refrain from formal hypothesis testing and instead report descriptive comparisons together with uncertainty estimates for our own pipeline.

Following the observation of particularly large relative gains for two classes with very few available training examples, we conducted a post-hoc exploratory analysis to assess whether official training-partition class size was associated with mean per-category F1 or with per-category $\deltaf$ relative to the published ConvNeXt point estimates, using Spearman rank correlations.

\section*{Results}
\label{sec:results}

Unless otherwise stated, all results below are reported on the official AQUA20 test set. A separate external validation of the same pipeline on the Sea Animals dataset, reported in the Supplementary Information, shows comparable performance and shape of the label-efficiency curve on a second, related image distribution.

\paragraph{Reference Numbers}
The published ConvNeXt point estimates are 88.9\% macro F1, 90.7\% accuracy, and 90.6\% support-weighted F1 (the latter reconstructed from its published per-category F1 scores using the official test-set supports)~\cite{fuad2026aqua20}. Note once again that the authors report point estimates of their model's performance.

\subsection*{Label Efficiency Across Annotation Budgets}

\begin{figure}[p]
    \centering
    \includegraphics[width=\columnwidth]{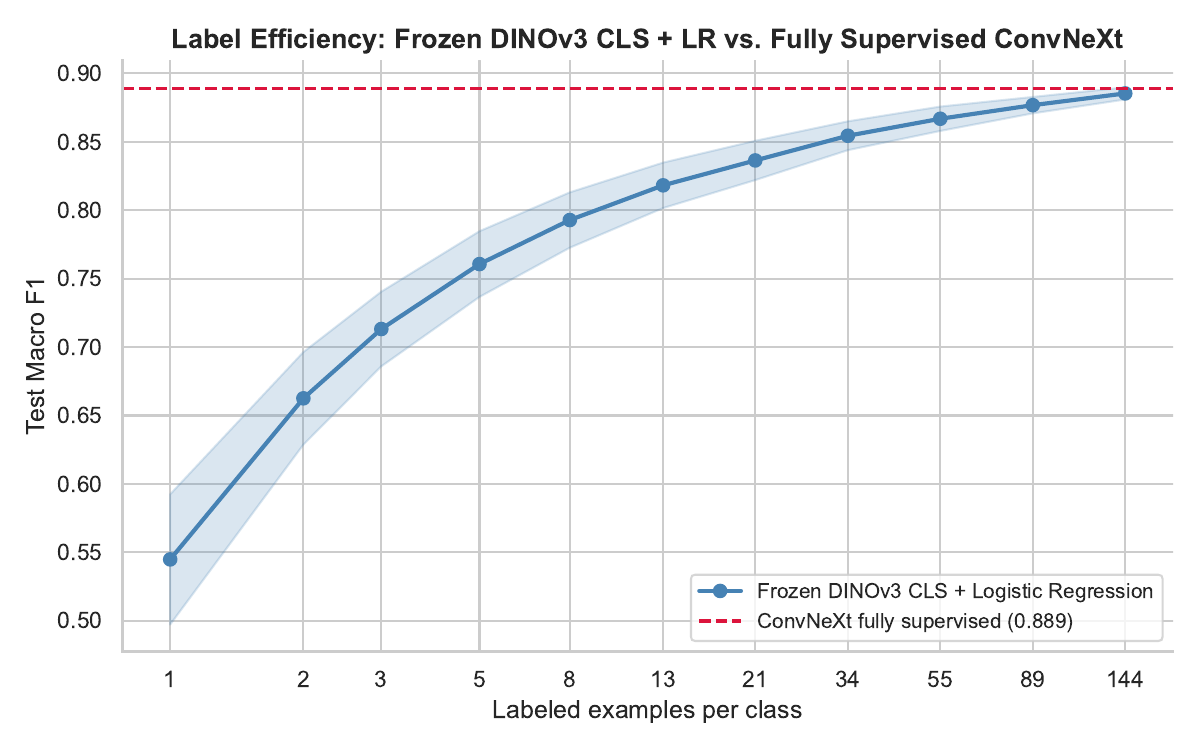}
\caption{Mean test-set macro F1 ($\pm$ one s.d.; $n = 100$ runs) across absolute annotation budgets of 1--144 labelled images per category. The published fully supervised ConvNeXt point estimate of 88.9\% macro F1~\cite{fuad2026aqua20} is shown as a dashed red line.}   \label{fig:budget_curve}
\end{figure}

Table~\ref{tab:budgets} summarises performance across the absolute annotation budgets, and Figure~\ref{fig:budget_curve} shows the corresponding learning curve. Performance increases steeply in the low-data regime before showing diminishing returns. With 13 labelled images per category, corresponding to 260 training images or approximately 4\% of the official training partition, mean macro F1 reaches 81.8\%. At a nominal cap of 89 images per category, corresponding to 1{,}305 actual training images after class-size capping and roughly 20\% of the official train data, mean macro F1 reaches 87.7\%. At 144 images per category, corresponding to 1{,}779 training images, mean macro F1 reaches 88.5\%. Relative to the published ConvNeXt macro F1, 2 of the 100 runs exceeded the reference at the 89-image budget and 20 of the 100 runs exceeded it at the 144-image budget; at the 55-image budget, 1 run exceeded the reference, while below that budget none did.

\begin{table}[t]
\centering
\caption{Mean test-set performance ($\pm$ one s.d.; $n = 100$ runs) across absolute annotation budgets. F1 is macro-averaged, while accuracy is calculated over all test observations. The published fully supervised ConvNeXt point estimates are 88.9\% macro F1 and 90.7\% accuracy. The rightmost column reports the percentage of individual runs whose macro F1 exceeded the published ConvNeXt point estimate. For the absolute-budget and 80/20 rows, the $\pm$ intervals denote one standard deviation across independent training-data samples. The 100\% refit is a deterministic point estimate; its bracketed interval is a 95\% class-stratified bootstrap confidence interval over the test set.}
\label{tab:budgets}
\begin{tabular}{lccr}
\toprule
Budget & Test F1 & Test Acc. & \% $>$ CX \\
\midrule
1/cls   & $0.545 \pm 0.048$ & $0.562 \pm 0.081$ & 0 \\
2/cls   & $0.663 \pm 0.034$ & $0.685 \pm 0.061$ & 0 \\
3/cls   & $0.713 \pm 0.027$ & $0.735 \pm 0.051$ & 0 \\
5/cls   & $0.761 \pm 0.024$ & $0.782 \pm 0.036$ & 0 \\
8/cls   & $0.793 \pm 0.020$ & $0.812 \pm 0.027$ & 0 \\
13/cls  & $0.818 \pm 0.017$ & $0.835 \pm 0.019$ & 0 \\
21/cls  & $0.836 \pm 0.014$ & $0.852 \pm 0.015$ & 0 \\
34/cls  & $0.854 \pm 0.011$ & $0.864 \pm 0.011$ & 0 \\
55/cls  & $0.867 \pm 0.009$ & $0.874 \pm 0.009$ & 1 \\
89/cls  & $0.877 \pm 0.006$ & $0.881 \pm 0.006$ & 2 \\
144/cls & $0.885 \pm 0.004$ & $0.887 \pm 0.005$ & 20 \\
\midrule
80/20   & $0.909 \pm 0.006$ & $0.914 \pm 0.003$ & 100 \\
100\%   & $0.915\;[0.890\text{--}0.937]$ & $0.918\;[0.904\text{--}0.931]$ & --- \\
\bottomrule
\end{tabular}
\end{table}

\subsection*{Repeated 80/20 Evaluation}
Across 100 repeated stratified 80/20 training-subsample runs, logistic regression on frozen DINOv3 embeddings achieves 90.9\% macro F1 ($\pm$ 0.6\% training-subsample s.d.; 95\% percentile subsample interval 89.6\%--92.2\%), 91.5\% support-weighted F1 ($\pm$ 0.3\%; 91.0\%--92.0\%), and 91.4\% accuracy ($\pm$ 0.3\%; 90.9\%--92.0\%) on the official test set.

All 100 training-subsample runs produced macro F1 values above the published ConvNeXt point estimate. 
Because AQUA20 reports only a single ConvNeXt result without run-to-run variability, this 100/100 count describes the position of our observed training-subsample distribution relative to a fixed reference; it does not estimate the probability of outperforming a newly trained ConvNeXt model.

\subsection*{Full-Training Performance}
Using 100\% of the official AQUA20 training partition with the hyperparameters selected in the separate validation procedure, logistic regression on frozen DINOv3 embeddings achieves 91.5\% macro F1, 91.9\% support-weighted F1, and 91.8\% accuracy on the official test set. Class-stratified bootstrapping of the test observations yields 95\% confidence intervals of 89.0--93.7\% for macro F1, 90.6--93.2\% for weighted F1, and 90.4--93.1\% for accuracy.

Because the fitted classifier and training data are fixed, the reported confidence intervals quantify uncertainty arising from the finite official test set rather than variability across training runs. 

\subsection*{Per-Class Analysis}
Analysing per-category performance across the absolute budgets reveals that the frozen embeddings become effectively comparable to the fully supervised ConvNeXt baseline on specific categories even under strict label scarcity. At a budget of just 8 labelled images per category, Octopus scores $0.762$ against ConvNeXt's $0.750$ fully supervised performance. At the same budget, MarineDolphin ($0.739$ vs.\ $0.737$) can also be favourably compared to the respective published baseline, with Crab ($0.858$ vs.\ $0.857$) following at 13 images per category. Full per-category F1 scores across all absolute annotation budgets are reported in the Supplementary Information.

\begin{figure}[p]
    \centering
    \includegraphics[width=\columnwidth]{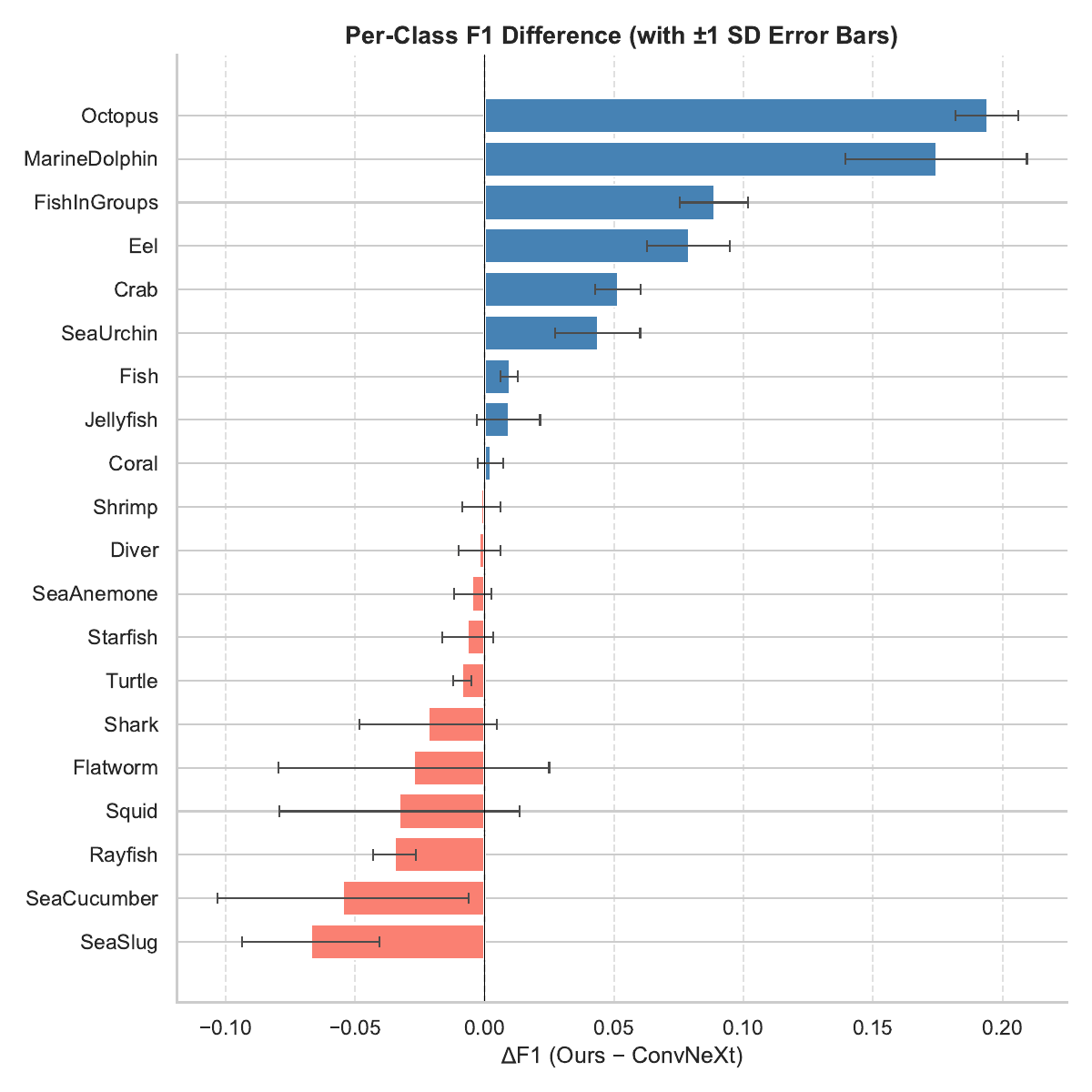}
    \caption{Difference in per-category F1 between logistic regression on frozen DINOv3 embeddings in the repeated 80/20 training-subsample evaluation ($n = 100$ runs) and the published ConvNeXt point estimates~\cite{fuad2026aqua20}. Differences are calculated as our mean F1 minus the ConvNeXt F1. Blue bars indicate categories for which our mean is higher; red bars indicate categories for which the published ConvNeXt value is higher.}
    \label{fig:delta_f1}
\end{figure}

In the repeated 80/20 evaluation, mean per-category F1 is higher than the published ConvNeXt value for 9 of the 20 categories. Among categories with larger test sets, the strongest gains occur for FishInGroups ($\deltaf = +0.089$; $n = 72$), Eel ($+0.079$; $n = 41$), and SeaUrchin ($+0.044$; $n = 29$), while Crab also improves by $+0.051$, although its test set contains only 11 images. The largest absolute gains occur for Octopus ($+0.194$) and MarineDolphin ($+0.174$), but both categories contain only 10 test images, so individual misclassifications have a large effect on their F1 estimates. The largest deficits relative to ConvNeXt are observed for SeaSlug ($-0.067$), SeaCucumber ($-0.055$), and Rayfish ($-0.035$). Table~\ref{tab:perclass} reports the complete comparison, and Figure~\ref{fig:delta_f1} visualises the differences.

A post-hoc Spearman analysis showed no meaningful monotonic association between training-partition class size and either mean per-category F1 (Spearman $\rho = 0.063$, $p = 0.791$) or per-category $\deltaf$ relative to ConvNeXt ($\rho = -0.080$, $p = 0.736$). These relationships can be examined in Supplementary Figures S3 and S4. The particularly large relative gains observed for the two classes with the fewest available training examples therefore do not appear to reflect a general tendency for smaller classes to favour the frozen-embedding approach.

\begin{table}[t]
\centering
\caption{Per-category F1 in the repeated 80/20 training-subsample evaluation ($n = 100$ runs) compared with the published ConvNeXt point estimates~\cite{fuad2026aqua20}. Values for our method are reported as mean $\pm$ one s.d. Bold $\deltaf$ values indicate categories for which our mean F1 is higher than the published ConvNeXt value; ties are not bolded. $n$ denotes the number of test-set observations.}
\label{tab:perclass}
\small
\begin{tabular}{l r cc r}
\toprule
Category & $n$ & Ours F1 & ConvNeXt F1 & $\deltaf$ \\
\midrule
Coral           & 348 & $0.908 \pm 0.005$ & 0.906 & $\mathbf{+0.002}$ \\
Crab            &  11 & $0.909 \pm 0.009$ & 0.857 & $\mathbf{+0.051}$ \\
Diver           &  13 & $0.998 \pm 0.008$ & 1.000 & $-0.002$ \\
Eel             &  41 & $0.914 \pm 0.016$ & 0.835 & $\mathbf{+0.079}$ \\
Fish            & 538 & $0.928 \pm 0.003$ & 0.918 & $\mathbf{+0.010}$ \\
FishInGroups    &  72 & $0.835 \pm 0.013$ & 0.746 & $\mathbf{+0.089}$ \\
Flatworm        &  13 & $0.819 \pm 0.052$ & 0.846 & $-0.027$ \\
Jellyfish       &  25 & $0.971 \pm 0.012$ & 0.962 & $\mathbf{+0.009}$ \\
MarineDolphin &  10 & $0.911 \pm 0.035$ & 0.737 & $\mathbf{+0.174}$ \\
Octopus         &  10 & $0.944 \pm 0.012$ & 0.750 & $\mathbf{+0.194}$ \\
Rayfish         &  95 & $0.929 \pm 0.008$ & 0.963 & $-0.035$ \\
SeaAnemone      & 221 & $0.894 \pm 0.007$ & 0.899 & $-0.005$ \\
SeaCucumber     &  10 & $0.893 \pm 0.048$ & 0.947 & $-0.055$ \\
SeaSlug         &  20 & $0.856 \pm 0.027$ & 0.923 & $-0.067$ \\
SeaUrchin       &  29 & $0.882 \pm 0.016$ & 0.839 & $\mathbf{+0.044}$ \\
Shark           &  19 & $0.843 \pm 0.026$ & 0.865 & $-0.022$ \\
Shrimp          &  11 & $0.951 \pm 0.007$ & 0.952 & $-0.001$ \\
Squid           &  10 & $0.856 \pm 0.046$ & 0.889 & $-0.033$ \\
Starfish        &  40 & $0.955 \pm 0.010$ & 0.962 & $-0.007$ \\
Turtle          &  76 & $0.978 \pm 0.004$ & 0.987 & $-0.009$ \\
\midrule
Macro avg.      &     & $0.909 \pm 0.006$ & 0.889 & $\mathbf{+0.020}$ \\
\bottomrule
\end{tabular}
\end{table}

\subsection*{Confusion Structure}

\begin{figure*}[p]
    \centering
    \includegraphics[width=0.90\textwidth]{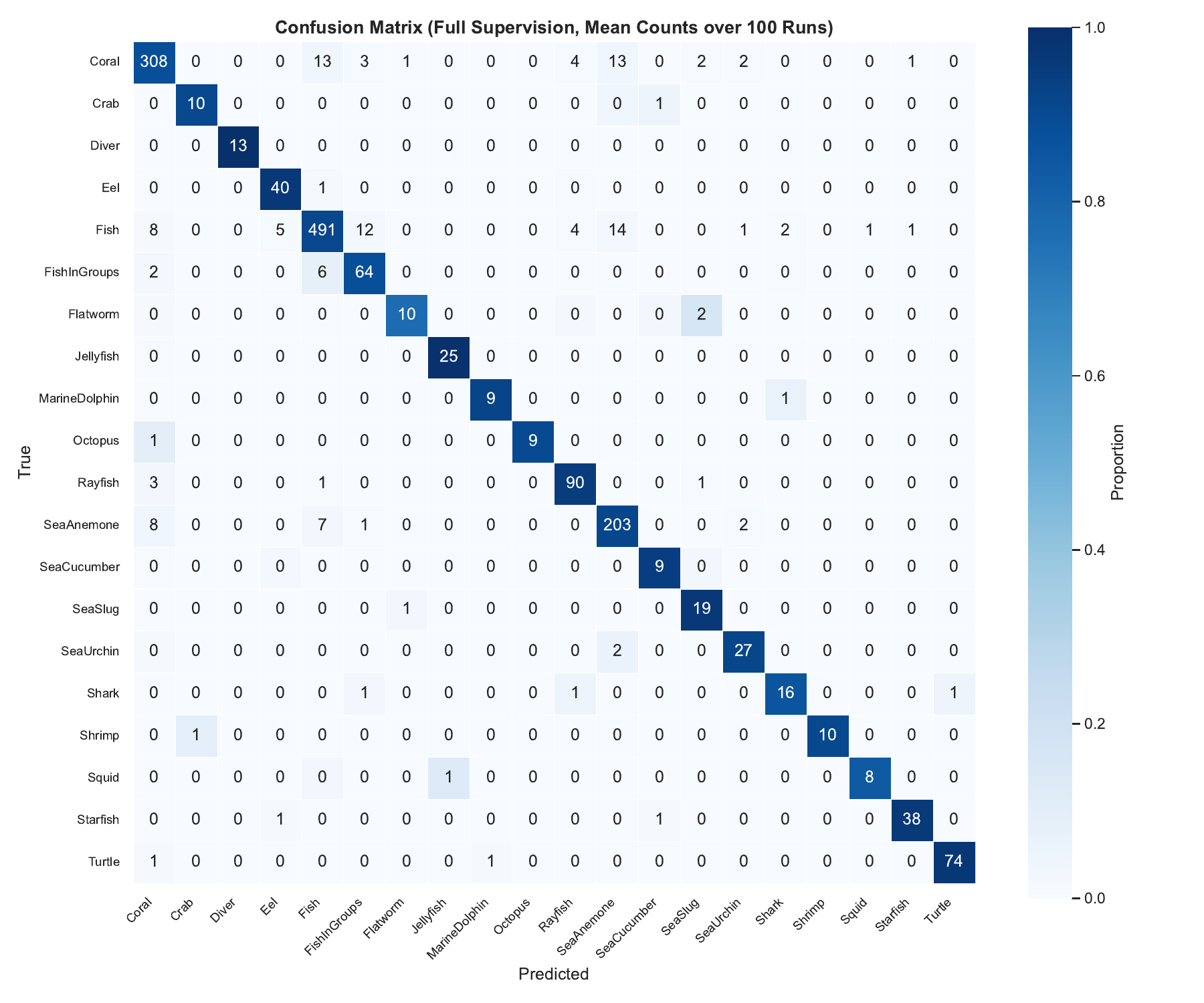}
    \caption{Confusion matrix for the repeated 80/20 training-subsample evaluation, showing mean counts over 100 runs and colours corresponding to row-normalised proportions.}
    \label{fig:confusion}
\end{figure*}

Figure~\ref{fig:confusion} shows the mean confusion matrix across the 100 repeated 80/20 training-subsample runs. Given the strong class imbalance in AQUA20, we primarily interpret the confusion matrix using row-normalised error rates rather than raw counts, which are heavily influenced by class prevalence.

After normalising each row by the number of test observations in the corresponding category, FishInGroups being classified as Fish is one of the clearest recurring confusions, affecting 8.9\% of FishInGroups observations.

Confusion between Coral and SeaAnemone is approximately bidirectional, with 3.7\% of Coral observations classified as SeaAnemone and 3.6\% of SeaAnemone observations classified as Coral. The highest row-normalised rates overall occur for Flatworm classified as SeaSlug (16.2\%), Squid as Jellyfish (12.9\%), and Octopus as Coral (10.0\%). However, these categories contain only 13, 10, and 10 test observations, respectively, so their estimated rates are highly sensitive to individual errors.

\subsection*{Dataset Audit}

The duplicate audit of the official AQUA20 split identified 94 of the 1{,}612 test images (5.8\%) with at least one duplicate or near-duplicate counterpart in the training partition, including 16 pixel-identical train/test pairs. The overlap is concentrated rather than uniform: 17.1\% of Turtle test images are affected, whereas 14 of the 20 categories contain no affected test images. Re-scoring the same fitted classifiers after excluding the 94 affected test images, with the training data and fitted models left unchanged, reduced macro F1 by at most 0.06 percentage points across both evaluations. Accuracy decreased from 91.4\% to 91.1\% and from 91.8\% to 91.5\%, respectively. The audit also identified 23 duplicate clusters containing images filed under more than one category label, including five cross-split duplicate or near-duplicate pairs with conflicting training and test labels. The complete audit is reported in Supplementary Note~1.

The audit findings have been communicated to the original AQUA20 authors.

\subsection*{Runtime Reporting}
\label{sec:runtime}

On an Apple M4 MacBook Air with 16\,GB RAM, extraction of the $384 \times 384$ embeddings using a batch size of four took 798.5 seconds for the 6{,}559 training images and 209.5 seconds for the 1{,}612 test images. Extraction for the complete AQUA20 dataset therefore took 1{,}008.0 seconds, or approximately 16.8 minutes, corresponding to an overall mean of approximately 0.12 seconds per image. The 100 repeated 80/20 classification runs completed in 9 minutes 32 seconds, corresponding to 5.72 seconds per run. For runtime measurement, the deterministic 100\% training-data refit was repeated 100 times and completed in 14 minutes 51 seconds, corresponding to approximately 8.9 seconds per fit; only one such fit is used for the reported full-training result. On the NVIDIA RTX A6000, embedding extraction for the complete dataset took approximately 2 minutes.

\section*{Discussion}
\label{sec:discussion}

Our results demonstrate that logistic regression on frozen, general-purpose DINOv3 embeddings achieves competitive category recognition on the AQUA20 benchmark at remarkably low annotation budgets. With just 13 labelled examples per class, macro F1 already exceeds 80\%; at 89 labelled examples, with a stratified rough 20\% of the training data, it approaches the reported point estimate of the benchmark; and at 144 labelled examples, using less than 30\% of the training data, its performance is comparable to the published fully supervised ConvNeXt point estimates, for which run-to-run variability was not reported. Beyond the logistic regression classifier, no component is trained: the DINOv3 backbone remains frozen, and the classifier is fitted via convex optimisation rather than end-to-end fine-tuning. Together with the runtime results, this supports the central practical claim of the paper: the pipeline is usable with relatively minimal labelling effort on commodity hardware and does not require domain-specific data engineering or underwater-adapted models. Because the foundation model was trained on general internet imagery with no marine-specific focus, these results may represent a baseline for what is achievable with this paradigm. Absolute AQUA20 scores should nevertheless be interpreted against the residual annotation noise and train/test overlap documented in the Limitations and Supplementary Note~1.

\paragraph{Label and Training Efficiency}
This has practical implications for marine scientists. Because the approach requires only a handful of reference images and a standard logistic regression classifier, it lowers the barrier to entry for researchers who lack the specialised deep learning expertise or GPU infrastructure required to train custom architectures, or the number of labelled images usually necessary for the deployment of fully end-to-end trained deep learning models in their research. The proposed pipeline is immediately applicable to pre-cropped images for rapid biodiversity assessments from opportunistic footage, preliminary inventories at new survey sites, and annotation bootstrapping for larger supervised projects.

To contextualize these runtimes: a ConvNeXt fine-tune on a comparable marine image classification task has been reported at approximately 13 images per second on a single NVIDIA V100 data-centre-grade GPU~\cite{pastore2023plankton}. While not directly comparable due to differences in image size and dataset, this would place 100 epochs over the AQUA20 training partition on the order of 14 GPU-hours, not accounting for the slower NVIDIA P100 used by the original authors~\cite{fuad2026aqua20}. Our pipeline completed in under 27 minutes on a consumer laptop with no discrete GPU.

\paragraph{Relevance for existing yet unexplored datasets}
Marine research institutions often hold large volumes of curated underwater imagery and video that remain unanalysed because the cost of building supervised classifiers for each collection has been prohibitive. With the proposed pipeline, a researcher without GPU infrastructure and with less machine-learning tooling than end-to-end deep learning requires can obtain category-level classifications approaching supervised accuracy from a modest number of labelled reference images. Retrospective analysis of such archival footage therefore becomes feasible, particularly when combined with active learning strategies that target low-certainty categories for additional annotation.

\paragraph{Complementary performance to fully realised deep learning}
While our frozen-embedding pipeline reaches a favourable macro F1 of 91.5\% in the full-training refit, compared with the published 88.9\% point estimate of AQUA20, it is important to note that this comparison is not symmetric in upstream investment. The DINOv3 ViT-B/16 used here was distilled from a 7B-parameter teacher trained on 1.7 billion images, whereas the ConvNeXt baseline was pretrained on ImageNet (approximately 1.3 million images) before fine-tuning on AQUA20. The frozen-embedding approach therefore leverages roughly three orders of magnitude more pretraining data. The practical claim of this work, that a researcher utilising these technologies need not train any neural network, remains valid, but the scientific comparison should be understood as contrasting two different strategies for amortising large-scale visual pretraining, rather than as a simple-versus-complex dichotomy.

In addition, the per-category analysis reveals a complementary performance pattern: our combination of logistic regression and general-purpose foundation model excels on specific low-sample-count categories (MarineDolphin, Octopus, Crab), while ConvNeXt retains an advantage on others (SeaSlug, SeaCucumber, Rayfish). The fact that each method outperforms the other on different subsets of the data could indicate that the general frozen embedding space captures different discriminative features than those prioritised by task-specific, end-to-end supervision. This demonstrates that end-to-end deep learning and linear classification on foundation features occupy distinct methodological niches. Furthermore, this complementarity suggests potential for performance gains through ensemble approaches or per-category routing, even without modifying the underlying frozen representations.

Crucially, this approach complements rather than replaces existing supervised pipelines also with regard to different axes. Lightweight end-to-end deep learning architectures such as YOLO~\cite{redmon2016yolo,terven2023yolo_review,muksit2022yolofish} occupy a fundamentally different region of the cost-performance frontier: they offer real-time, on-device inference~\cite{hampau2022edge}, essential for tasks like autonomous vehicle navigation or live camera monitoring, but exact a high upfront cost to a project by requiring large, exhaustively labelled datasets, infrastructure, and scarce knowledge for training. The foundation model paradigm presented here, however, excels at the offline analysis of existing imagery, enabling rapid, label-efficient processing of historical data where exhaustively annotating a new training set is economically infeasible.

\paragraph{Class imbalance}
AQUA20 exhibits severe class imbalance, with per-category sample counts ranging from 30 (MarineDolphin, Octopus) to 2,738 (Fish). This has distinct consequences for model fitting and for the stability of per-category evaluation. First, to counteract the imbalance during fitting, the logistic regression classifier uses balanced class weights, which up-weight rare categories in proportion to their underrepresentation. The small difference between macro F1 and support-weighted F1 in both the repeated 80/20 evaluation (90.9\% vs. 91.5\%) and the 100\% refit (91.5\% vs. 91.9\%) indicates that aggregate performance was not driven primarily by the most prevalent categories. If rare classes were systematically underperforming, macro F1, which weights all categories equally, would fall substantially below the support-weighted variant. Second, the two largest per-category gains over ConvNeXt in the repeated 80/20 evaluation occur for Octopus ($\deltaf = +0.194$) and MarineDolphin ($\deltaf = +0.174$), which are also the two smallest classes in the training partition. To assess whether this reflects a systematic tendency for smaller classes to benefit more from the frozen-embedding approach, we conducted a post-hoc Spearman analysis, which found no monotonic association between training-partition class size and either per-category F1 ($\rho = 0.063$, $p = 0.791$) or per-category $\deltaf$ relative to ConvNeXt ($\rho = -0.080$, $p = 0.736$). The large gains for these two categories therefore appear to be category-specific rather than a general consequence of class size: Shrimp (22 training images) and Squid (24 training images) are comparably rare yet show no advantage over ConvNeXt ($\deltaf = -0.001$ and $-0.033$, respectively). Third, class imbalance limits the reliability of per-category evaluation for the rarest classes. Categories such as Flatworm, Squid, and Octopus contain only 10 to 13 test observations each, so individual misclassifications produce large fluctuations in estimated F1 and error rates, as reflected in the higher standard deviations and less stable confusion rates reported for these categories.

\paragraph{Confusion structure}
The bidirectional confusion between Coral and SeaAnemone, 3.7\% and 3.6\% per sample, is consistent with their shared anthozoan morphology~\cite{fuad2026aqua20}. The FishInGroups-to-Fish confusion rate of 8.9\% likely reflects the inherent ambiguity of a category boundary defined by the number of visible individuals rather than morphology. Some of this confusion is also consistent with residual annotation noise: as reported in the dataset audit, five cross-split duplicate or near-duplicate pairs carry conflicting train/test labels, all five within the Coral, SeaAnemone, and Fish cluster (Supplementary Note~1).

For some categories, a certain degree of confusion is in addition likely irreducible. Coral formations frequently host other organisms, and fish such as clownfish are routinely photographed within host anemones, making the assignment of a single ground-truth label inherently ambiguous in these cases. Several of the highest per-sample confusion rates involve rare classes with fewer than 15 test samples, where individual misclassifications have outsized effects on the estimated rate. In addition, biological adaptations such as some species of flatworms mimicking toxic sea slugs~\cite{newman1994flatworm_mimicry} highlight the inherent difficulty of this classification task: while the statistical analysis of these rare classes is unreliable, one might interpret that our classifier may in some cases simply fall victim to adaptations that evolved to deceive the natural visual neural networks of predators.

\subsection*{Limitations}\label{sec:limitations}

Several limitations should be noted. 
First, our primary evaluation uses a single dataset (AQUA20) and a single foundation model (DINOv3 ViT-B/16). An external validation on the Sea Animals dataset (Supplementary Information) provides evidence that the pipeline and learning-curve shape transfer to a second image distribution, but that dataset is not actively curated, has not previously been officially benchmarked for full deep learning approaches, carries an unknown amount of underwater degradation and is approximately class-balanced, so generalisation across challenging underwater environments, severe class imbalance, and alternative embedding models remains to be conclusively established. In addition, we neither compare DINOv3 against domain-adapted models such as AquaticCLIP~\cite{alawode2025aquaticclip} nor against its predecessor, DINOv2; our selection is based on proxy evidence from previous marine machine learning literature on DINOv2~\cite{diveseg2025}, terrestrial species clustering~\cite{markoff2026vit_clustering}, and the deliberate decision to evaluate only general-purpose, off-the-shelf representations.

Although it does not replace direct evaluation within our experimental protocol, the concurrent work of Saleh and Rahimi Azghadi provides relevant converging evidence. Their study compares DINOv2 linear probing with LoRA, VPT, and full fine-tuning across five marine datasets, including a subset of AQUA20, with evaluations spanning low-label settings~\cite{saleh2026howmanylabels}. Its results suggest that the broader pattern reported here is not confined to DINOv3 or to the two datasets evaluated in our study.

A second limitation concerns the comparison with ConvNeXt. AQUA20 reports only a single published point estimate for the baseline and does not provide run-to-run variability. Retraining ConvNeXt under our repeated-subsample protocol would require independently reimplementing the published training and evaluation pipeline, introducing additional implementation choices that could materially affect the resulting comparison. We therefore treat the published ConvNeXt result as a fixed reference and report training-sample variability only for our own method.

Third, because AQUA20 contains pre-cropped images and defines a closed set of 20 known categories, our evaluation does not address localisation in raw imagery, background-only frames, multiple organisms, or previously unseen categories.
Applying the pipeline to unprocessed imagery or video would therefore require an upstream detection, segmentation, or cropping stage, as well as mechanisms for handling unknown categories. Recent advances in zero-shot concept segmentation, notably SAM~3~\cite{carion2025sam3}, which returns instance masks for all occurrences of a text-described concept without task-specific training and whose evaluation benchmark includes underwater imagery from FathomNet~\cite{katija2022fathomnet}, indicate that such upstream stages need not reintroduce the complexity this pipeline avoids; the present comparison nonetheless evaluates the classification stage in isolation and does not measure end-to-end performance of a complete pipeline.

Fourth, while we report ablations over input resolution, alternative classifiers, regularisation sensitivity, pooling strategy, solver convergence, and dimensionality reduction (all in the Supplementary Information), their scope is limited. No hyperparameter search was conducted per absolute budget, and more extensive exploration of kernels, calibrated classifiers, ensembles, or semi-supervised methods may yield further improvements. For researchers with preinvestment in deep learning hardware, either choosing the original 7B DINOv3 model or an even larger input resolution than our selected 384×384 may potentially yield even better performance.

Fifth, the duplicate audit reported above and in Supplementary Note~1 indicates that AQUA20 contains residual train/test overlap and annotation noise. This places a mild optimistic bias on absolute performance estimates while also imposing an aleatoric ceiling on the maximum attainable score. We consciously retain the official split as our main reporting target even accounting for the audit findings to preserve comparability with the published ConvNeXt baseline. Because re-scoring on the duplicate-filtered test subset reduced our macro F1 only slightly, the overlap is unlikely to account for the reported margin over ConvNeXt. It should be noted, however, that duplicated test images might be recognisable through memorisation~\cite{barz2020cifar}, and an end-to-end ConvNeXt trained for 100 epochs has greater capacity to memorise individual training images than a linear classifier on frozen, label-agnostic features. It is therefore reasonable to assume that the published baseline benefits at least as much as our method from such overlap. Under the conservative assumption that removing duplicates would not increase the baseline score, the 80/20 margin remains at least 1.9 percentage points.

The audit also identified duplicate clusters containing images filed under more than one class label, including five cross-split duplicate or near-duplicate pairs with conflicting train/test labels. These cases indicate residual taxonomic or scene-level ambiguity in AQUA20 and are consistent with the confusion structure observed among Coral, SeaAnemone, Fish, and FishInGroups. They also imply that some high-end errors are not purely algorithmic: a classifier may be trained on one label for a near-identical image and evaluated against another.

Sixth, we cannot exclude overlap between AQUA20 imagery and the web-scale corpus (LVD-1689M) used to pretrain the teacher 7B DINOv3 which subsequently trained the 86M model used in our pipeline. Some AQUA20 images were reportedly drawn from the EUVP dataset and other publicly available sources~\cite{fuad2026aqua20}. EUVP was assembled from oceanic-exploration and human--robot field footage captured with several cameras, supplemented with frames extracted from publicly available online videos~\cite{islam2020euvp}; this web-sourced component in particular cannot be assumed disjoint from a large corpus such as LVD-1689M while it is not reported to be included in the original ImageNet dataset. Any such overlap would benefit the frozen-embedding approach but not the ImageNet-pretrained baseline, and it is inherent to comparing a web-pretrained foundation model against a smaller-corpus baseline. Because DINOv3 pretraining is fully self-supervised, and the ViT-B/16 checkpoint used here is itself distilled from a larger teacher, any such overlap could improve the quality of the frozen representation on similar images but cannot transfer category labels; we flag this as an unquantifiable bound rather than a measured effect.

\subsection*{Future Work}

Although the Sea Animals validation (Supplementary Information) provides initial evidence of cross-dataset transfer, evaluation on additional underwater-specific datasets with realistic degradation conditions is necessary to establish the generality of our findings. Further work should compare DINOv3 with other general-purpose and marine-adapted foundation models and evaluate the detection, segmentation, or cropping stages required to apply the classifier to unprocessed imagery and video. Naturally, the potential of a DINOv3-derived fine-tune specifically trained on underwater imagery could be established against the base model. Finally, the complementary per-category strengths observed between logistic regression and ConvNeXt motivate investigation of ensemble or method-selection strategies, while calibrated uncertainty and active learning may help prioritise additional annotation.

\section*{Conclusion}
\label{sec:conclusion}
We have shown that frozen DINOv3 embeddings provide a strong basis for label-efficient underwater image classification using a simple logistic regression classifier. With 13 labelled images per category, corresponding to 260 images or approximately 4\% of the official training partition, macro F1 reaches 81.8\%; with 89 and 144 images per category, it reaches 87.7\% and 88.5\%, respectively, compared to the published fully supervised ConvNeXt point estimate of 88.9\%. Performance continues to improve as more labelled data are used, reaching 90.9\% macro F1 in the repeated 80/20 training-subsample evaluation and 91.5\% after refitting on the complete official training partition. Preliminary external validation on the Sea Animals dataset further suggests that the general learning-curve behaviour is not unique to AQUA20. Because the DINOv3 backbone remains frozen and only the downstream classifier is fitted, the approach avoids task-specific neural-network training and can be executed on commodity hardware. These findings indicate that much of the visual structure required for underwater category recognition is already present in the frozen representation, making linear classification on foundation-model embeddings a practical baseline where annotation effort, computational resources, or machine-learning expertise are limited.

\section*{Data availability}
The AQUA20 benchmark dataset used in this study was introduced
by Fuad et al.~\cite{fuad2026aqua20} and is publicly available
from the original authors. The Sea Animals dataset used for external validation
is publicly available from Kaggle~\cite{seaanimals_kaggle} (please note: the original URL says \texttt{dataste}, it is not a typographical error); source images are
not redistributed. The embedding and classification code is available at
\url{https://github.com/MiaSpugnetta/label-efficient-underwater-image-clf}.
Pre-computed embeddings for both
AQUA20 and Sea Animals are archived on Zenodo at
\url{https://zenodo.org/records/20853142}.

\bibliography{refs}

\section*{Funding}
This study was supported by the EU grants TWILIGHTED (Horizon Europe Grant No. 101158714) and SEAMPHONI (Horizon Europe Grant No. 101206245); by Portuguese national funding for Associate Laboratory MARE (FCT Grant No. UIDB/04292/2020) and Associate Laboratory ARNET (FCT Grant No. LA/P/0069/2020); and by FCT projects 10.54499/LA/P/0083/2020 and UID/50009/2025.

\section*{Use of generative AI tools}
During the preparation of this work, the authors used AI-assisted tools, including ChatGPT (GPT-5.5), Claude (Opus 4.6), and Gemini (3.1 Pro), to support code development, debugging, and manuscript editing. These tools were not used to define the research question, select the dataset, make study-design decisions, conduct final analyses, interpret the results, or draw scientific conclusions. All AI-assisted outputs were reviewed, edited, and checked by the human authors against the underlying code, data, and numerical results before inclusion. The authors take full responsibility for the content of the published article.

\section*{Author contributions statement}
\textbf{Thomas Manuel Rost:} Conceptualization, Data curation, Formal analysis, Investigation, Methodology, Resources, Software, Validation, Visualization, Writing of original draft.
\textbf{Martina Figlia:} Validation, Visualization, Software – review \& editing, Writing – review \& editing.
\textbf{F. Morgado-Dias:} Writing – review \& editing.
\textbf{Marko Radeta:} Investigation, Supervision, Validation, Writing – review \& editing.

\section*{Competing interests}
Thomas Manuel Rost and Martina Figlia provide consulting services and
commercial products based on foundation model classification pipelines
related to the approach described in this work. The remaining authors
declare that they have no known competing financial interests or personal
relationships that could have appeared to influence the work reported in
this paper. The selection of the published ConvNeXt benchmark as the reference baseline, the foundation model and classifier, all hyperparameters, and the evaluation protocols were determined independently of any commercial engagement. All components of our pipeline are available as open source, no component is derived from any commercial product. 

\end{document}